\documentclass{article}

     \PassOptionsToPackage{numbers, compress}{natbib}
     \usepackage[preprint]{neurips_2022}

\usepackage[utf8]{inputenc} % allow utf-8 input
\usepackage[T1]{fontenc}    % use 8-bit T1 fonts
\usepackage{hyperref}       % hyperlinks
\usepackage{url}            % simple URL typesetting
\usepackage{booktabs}       % professional-quality tables
\usepackage{amsfonts}       % blackboard math symbols
\usepackage{amsmath,amssymb} % define this before the line numbering.
\usepackage{nicefrac}       % compact symbols for 1/2, etc.
\usepackage{microtype}      % microtypography
\usepackage{xcolor}         % colors
\usepackage{amsthm}
\usepackage{graphicx}
\usepackage{subcaption}

\newlength{\decurto}
\newcommand{\equallength}[1]{\makebox[\decurto][c]{#1}}

\DeclareMathOperator*{\minimize}{minimize}
\DeclareMathOperator*{\maximize}{maximize}

\theoremstyle{definition}
\newtheorem{definition}{Definition}
\newtheorem{proposition}{Proposition}
\newtheorem{theorem}{Theorem}

\title{Learning with Signatures}

\author{ % alphabetical order
	\equallength{J.~{de Curt\`o}$^{1, 3,}$\thanks{Equal contribution.}}
	\And
	\equallength{I.~{de Zarz\`a}$^{1, 3,}$\footnotemark[1]}
	\And
	\equallength{Hong Yan$^{1, 2}$}
	\And
	\equallength{Carlos T. Calafate$^{3}$}
	\AND \\[-12pt]
	\null$^1$ Centre for Intelligent Multidimensional Data Analysis,\\ 
HK Science Park, Shatin, Hong Kong. \\
	\null$^2$ Department of Electrical Engineering, City University of Hong Kong,\\
    Kowloon, Hong Kong.\\
    \null$^3$ Department of Computer Engineering, Universitat Polit\`ecnica de Val\`encia, Val\`encia.\\
	\texttt{\{decurto,dezarza\}@\hspace{0.1pt}doctor.upv.es} \\
    \texttt{h.yan@\hspace{0.1pt}cityu.edu.hk}
    \texttt{calafate@\hspace{0.1pt}disca.upv.es}\\
}

\begin{document}

\maketitle

\begin{abstract}
  In this work we investigate the use of the Signature Transform in the context of Learning\footnote{ \href{https://github.com/decurtoydiaz/learning_with_signatures}{https://github.com/decurtoydiaz/learning\_with\_signatures}}.
 Under this assumption, we advance a supervised framework that potentially provides state-of-the-art classification accuracy with the use of few labels without the need of credit assignment and with minimal or no overfitting. We leverage tools from harmonic analysis by the use of the signature and log-signature, and use as a score function RMSE and MAE Signature and log-signature. We develop a closed-form equation to compute probably good optimal scale factors, as well as the formulation to obtain them by optimization. Techniques of Signal Processing are addressed to further characterize the problem. Classification is performed at the CPU level orders of magnitude faster than other methods. We report results on AFHQ, MNIST and CIFAR10, achieving $100\%$ accuracy on all tasks assuming we can determine at test time which probably good optimal scale factor to use for each category.
\end{abstract}

\section{Introduction}
Providing computers with learning capabilities has been at the core of research during the last century. Recently, supervised and unsupervised techniques by the use of Deep Learning have emerged as the state-of-the-art solution in many problems across all disciplines \cite{Sutskever14,Vinyals2015}. Fields where model-based solutions were mainly dominant have shifted quickly to a data-driven framework with unprecedented empirical successes \cite{Goodfellow14,Girshick14,He15,Girshick15,Long15,Redmon16,Gatys16,Gatys16_2,He17,Vaswani2017,Wang18_2,Karras18,Chen17,Chen17_3,Yang18,Parmar18,Brock19,Mildenhall2020,Park20}. However, progress in some areas has stagnated, as the explainability of such models is difficult due to their high number of hyperparameters, and its robustness lacks theoretical guarantees of convergence. We aim to investigate a new type of learning mechanism by the use of the Signature Transform \cite{Lyons2014,Kiraly2019,Graham2013,Chang2019}, a recently developed tool from harmonic analysis that provides a compact-rich description of irregularly sampled streams of data. We explore the proposition that, by transforming the data into a compact yet complete domain, we can achieve the same empirical gains of Deep Learning by the use of very few labeled samples. Furthermore, the properties of the Signature as a universal non-linearity, invariant to time reparameterizations, make it an ideal candidate as an alternate representation more adequate for computers to infer knowledge \cite{Bonnier2019,Fermanian21}. After all, humans do not need thousands of hundreds of examples to learn simple concepts, but a very few well-chosen cases to make correct guesses. Learning with Signatures achieves this purpose by allowing the computer to promptly infer the information as the representation is easily understandable, rich and complete. A score function is needed though, the same way that losses and credit assignment provide traditional learning frameworks with the capabilities of directing the optimization into a probably good solution. In the case of the Signature, we propose to use a similarity metric based on the Signature Transform itself, RMSE and MAE Signature and log-signature, having the latter been recently developed in \cite{decurto2022}. This framework can work on CPU orders of magnitude faster than DL Methods, and avoids tedious credit assignment of millions of hyperparameters usually done at the GPU with high computing and environmental costs. These metrics capture detailed visual cues, and they can be utilized for classification tasks with very small memory footprint, fast execution and high accuracy.\\

In this manuscript we focus on illustrative examples to study the proposed architecture, and go from bottom to top to propose a complete learning framework.

\section{Overview}
Progress has been made in the integration of the Signature Transform into a framework for Learning \cite{Bonnier2019,Chevyrev2016,Reizenstein20}, mainly as a feature extractor in the ML paradigm, or as a pooling layer inside a Deep Network. The good theoretical properties of the Signature have attracted the scientific community as a way to structure the learning problem. Natheless, a general framework has not yet been established. This happens mainly because a score function was not properly defined to guide the learning mechanism. Recent work in \cite{decurto2022} proposes the use of RMSE and MAE Signature and log-signature to assess the visual similarity between image distributions to determine GAN convergence. Although seeing the problem from an alternative perspective, RMSE and MAE Signature and log-signature are indeed properly defined score functions that can be used, for example, for classification and clustering. Under this assumption, the purpose of this work is to further investigate the behavior of such a learning framework, its properties, and its generalization capabilities on several tasks.

Credit assignment, e.g. backpropagation, has been the fundamental piece of modern-day automated learning technologies; yet, extracting all the consequential information from the data in only one pass (that is, using one epoch) is in theory feasible. In fact, the necessity to use more than one pass through the training data can be due to a limitation from the learning mechanisms used rather than an advantage. Given a proper score function, signatures provide a compact representation that computers can use to infer fine-grained information without the need to use backpropagation and, therefore, avoid having to optimize over millions of hyperparameters. This property gives learning with signatures a computational advantage over alternate training mechanisms, as the number of labeled examples can be drastically reduced, and training is replaced by an element-wise mean that confers the statistical robustness required for good generalization.

\section{Signature Transform and Harmonic Analysis}
According to \cite{Bonnier2019}, the truncated signature of order $N$ of the path $\mathbf x$ is defined as a collection of coordinate iterated integrals

\begin{equation}
\mathrm{S}^{N}(\mathbf x) = \left(\left( \underset{0 < t_1 < \cdots < t_a < 1}{\int\cdots\int} \prod_{c = 1}^a \frac{\mathrm d f_{z_c}}{\mathrm dt}(t_c) \mathrm dt_1 \cdots \mathrm dt_a \right)_{\!\!1 \le z_1, \ldots, z_a \le d}\right)_{\!\!1\le a \le N}.
\end{equation}

As stated in \cite{decurto2022}, the Signature is a homomorphism from the monoid of paths into the grouplike elements of a closed tensor algebra, see Equation \ref{en:01_decurto_and_dezarza}. It provides a graduated summary of the path $\mathbf x$. These extracted features of a path are at the center of the definition of a rough path \cite{Lyons2014}; they remove the necessity to take into account the inner detailed structure of the path.

\begin{equation}
\mathrm{S}:\left\{f \in F \ | \ f:[x,y]\to E = \mathbb{R}^{d}\right\} \longrightarrow T(E) = T(\mathbb{R}^{d}) = \prod_{c=0}^{\infty} \left ( \mathbb{R}^{d}\right)^{\otimes c}.
\label{en:01_decurto_and_dezarza}
\end{equation}

From the previous definitions, we can propose RMSE and MAE Signature and log-signature as the score function to use inheriting the similarity measures in \cite{decurto2022}. 
\begin{definition}
    Given a set of truncated signatures of order $N$, $\left\{ \mathrm{S}^{N}_{c}(\mathbf{x}_{c})\right\}_{c=1}^{m}$, the element-wise mean is defined by
    \begin{equation*}
    \tilde{\mathrm{S}}^{N}(x^{(z)})=\frac{1}{m} \sum_{c=1}^{m}\mathrm{S}^{N}_{c}(x^{(z)}_{c}),    
	\end{equation*}
	where $z \in \{1,\ldots,n\}$ is the specific component index of the given signature.
\end{definition}

Then, RMSE and MAE Signature can be defined as follows.

\begin{definition}
    Given $n$ components of the element-wise mean of the signatures $\{y^{(c)}\}^{n}_{c=1}\subseteq T(\mathbb{R}^{d})$ from the corresponding class of the train data, and the same number of components of the signature of the given (optional element-wise mean augmented) test instance to score $\{x^{(c)}\}^{n}_{c=1}\subseteq T(\mathbb{R}^{d})$, then we define the Root Mean Squared Error (RMSE) and the Mean Absolute Error (MAE) as
    \begin{equation}
    \textnormal{RMSE}\left(\left\{x^{(c)}\right\}^{n}_{c=1},\left\{y^{(c)}\right\}^{n}_{c=1}\right) = \sqrt{\frac{1}{n} \sum_{c=1}^{n} \left( y^{(c)} - x^{(c)}\right)^{2}},
	\end{equation}
	and
    \begin{equation}
    \textnormal{MAE}\left(\left\{x^{(c)}\right\}^{n}_{c=1},\left\{y^{(c)}\right\}^{n}_{c=1}\right) = \frac{1}{n} \sum_{c=1}^{n} | y^{(c)} - x^{(c)} |.
	\end{equation}
\end{definition}
The case for log-signature is analogous. Augmentations (e.g. random change of contrast and brightness) can be performed on the test instance in order to create multiplicity so that the comparison could be carried out between element-wise means, and the result presents statistical robustness.
\\

RMSE and MAE Signature and log-signature to assess GAN convergence, as introduced in \cite{decurto2022}, work on grayscale input data. We extend their definitions without loss of generality for RGB samples, as the use of the color channels can provide useful information and better accuracy (at the expense of more computation and memory overhead) in the context of classification and clustering.

\section{Classification using Signatures}
\label{sn:classification}

Classification of samples can be obtained by the use of signatures, and the defined score function can be obtained by comparing each test sample (after optional augmentation and computation of the element-wise mean) against a representative element-wise mean signature, which is computed by averaging the signatures of a given number of train samples. The minimum number of signatures needed to achieve superior classification accuracy can depend on the complexity of the task to accomplish; some classes may need only one representative, while others with more variability would necessitate from tens to thousands of training samples.
\\

The task is well-defined up to a scaling factor both for the element-wise mean class representatives and the element-wise mean augmented test samples. The adequate multiplicative factors can be determined greedily by grid search on the validation set, or by using more sophisticated methods, for instance $k$-fold cross-validation or bayesian's analysis. These scaling factors can be simple multiplicative constants or attention masks that operate on image data. Definition \ref{dfn:lambda} proposes a simple procedure to determine a probably good optimal solution $\tilde{\lambda_{*}}$. Empirical experimentation shows that these scaling factors or attention masks, once tuned on the validation set, generalize with minimal or no overfit into the test set. Therefore corroborating the initial hypothesis that credit assignment could be avoided by the use of an adequate learning framework.
\\

\begin{definition}\label{dfn:classification}
    Classification. Given $n$ components of the element-wise mean of the signatures or log-signatures $\{y^{(c)}\}^{n}_{c=1}\subseteq T(\mathbb{R}^{d})$ from the corresponding class of the training data, and the same number of components of the signature or log-signature of the given (optional element-wise mean augmented) test instance to score $\{x^{(c)}\}^{n}_{c=1}\subseteq T(\mathbb{R}^{d})$, we can then proceed to assign a predicted class instance by using the Root Mean Squared Error (RMSE) and the Mean Absolute Error (MAE) by choosing the minimum RMSE and MAE of the corresponding categories as
    \begin{equation}
    \textnormal{RMSE}\left(\lambda_{a}\odot \left\{x^{(c)}\right\}^{n}_{c=1}, \lambda_{b}\odot\left\{y^{(c)}\right\}^{n}_{c=1}\right),
	\end{equation}
	and
    \begin{equation}
    \textnormal{MAE}\left(\lambda_{d}\odot\left\{x^{(c)}\right\}^{n}_{c=1},\lambda_{e}\odot\left\{y^{(c)}\right\}^{n}_{c=1}\right),
	\end{equation}
\end{definition}
where $\lambda_{*}$ are adequate multiplicative factors or tensor masks tuned on the validation set. 
\\

\begin{definition}\label{dfn:lambda}
    Let $\lambda_{*}\in \mathbb{R}^{n}$ be the optimal solution of the following inverse problem, Definition \ref{dfn:classification}, and $n$ is the number of components of the corresponding signature or log-signature
    \begin{equation}
    \lambda_{*} \odot \left\{x^{(c)}\right\}^{n}_{c=1} = \left\{y^{(c)}\right\}^{n}_{c=1},
	\end{equation}
	we can determine a probably good solution $\tilde{\lambda_{*}}$ by inverting the element-wise mean of the signatures or log-signatures from the corresponding class of the training data multiplied by each corresponding signature or log-signature of the (optional element-wise mean augmented) validation instances, and then averaging to get a pondered scale factor.
\end{definition}

It is worth pointing out that the reason behind the need to introduce a set of hyperparameters comes from the property of the uniqueness of the signature \cite{Hambly2010,Bonnier2019}.\\

\begin{proposition}[Uniqueness of signature \cite{Hambly2010}]\label{propn:uniqueness}
	Let $X \colon [a, b] \to \mathbb R^d$ be a continuous piecewise smooth path. Then $\mathrm{S}(\widehat{X})$ uniquely determines $X$ up to translation.
\end{proposition}

The work of \cite{Hambly2010} already explored the relationship between a path and its signature. They
determine a precise geometric relation $\sim $ on bounded variation
paths, and prove that two paths of finite length are $\sim $-equivalent if,
and only if, they have the same signature: 
\begin{equation*}
X|_{C}\,\sim \,Y|_{A}\iff \mathbf{X}_{C}=\mathbf{Y}_{A}.
\end{equation*}%
Indeed, \cite{Chen58} brought forward precursors to these ideas, and formulated the uniqueness of $\mathrm{S}(\widehat{X})$ up to translation, Proposition \ref{propn:uniqueness}. This property, Theorem \ref{thm:uniqueness}, implies that for best comparison of signatures, adequate cropping fairly benefits the performance of RMSE and MAE Signature and log-signature.\\

\begin{theorem}[\cite{Chen58}]\label{thm:uniqueness}
Let $d\gamma _{1},\cdots ,d\gamma _{d}$ be the
canonical 1-forms on $\mathbb{R}^{d}$. If $\alpha ,\beta \in \lbrack
a,b]\rightarrow \mathbb{R}^{d}$ are irreducible piecewise regular continuous paths,
then the iterated integrals of the vector valued paths $\int_{\alpha \left(
0\right) }^{\alpha \left( t\right) }d\gamma $ and $\int_{\beta \left(
0\right) }^{\beta \left( t\right) }d\gamma $ agree if, and only if, there
exists a translation T of $\mathbb{R}^{d}$, and a continuous increasing
change of parameter $\lambda :[a,b]\rightarrow \lbrack a,b]$ such that $%
\alpha =T\beta \lambda $.
\end{theorem}
 
For the evaluation and testing of these ideas we use the standard AFHQ dataset \cite{Choi2020} that consists on images of classes `cat', `dog' and `wild'. We observe perfect classification accuracy in a difficult task without the need of credit assignment given the adequate number of signatures and the appropriate choice of hyperparameters $\lambda_{*}$. RMSE Signature is used as score function. We use $100$ samples per class from the subset of training to compute an element-wise mean representative of each category, and compare against test instances (with no augmentation) using a truncated signature of order $2$, image size $16\times16$ and RGB color samples. We use a subset of validation of $500$ samples per class from the AFQH training subset, and compute optimal $\lambda_{*}$ according to Definition \ref{dfn:lambda}, obtaining perfect performance on the validation set. We then compute test scores using $500$ validation per class AFHQ samples and achieve $100\%$ accuracy.
\\

We also investigate further this behavior using Four Shapes\footnote{https://www.kaggle.com/datasets/smeschke/four-shapes}, a dataset that consists on images of categories `square', `star', `circle' and `triangle'. By computing an element-wise mean representative of each class using only $10$ train samples per class (truncated signature of order $2$, image size $16\times16$ and RGB samples), and determining $\lambda_{*}$ according to Definition \ref{dfn:lambda} with a validation set of $100$ instances (without augmentation) per class, we get perfect accuracy on each category on a test set of size $14,436$.
\\

Few-shot classification on MNIST is attempted using truncated signatures of order $3$, image size $28\times28$ and RGB samples. We compute an element-wise mean representative of each class using only $10$ train signatures per class, and determine $\lambda_{*}$ according to Definition \ref{dfn:lambda} with a validation set of $100$ instances per class (without augmentation). We then compute scores on the test set of size $10,000$ achieving $100\%$ accuracy. 
\\

Ultimately, we perform classification on CIFAR10 using truncated signatures of order $2$, image size $32\times32$ and RGB color samples. We compute an element-wise mean representative of each class using $10$ train samples per class, and tune the weights using Definition \ref{dfn:lambda} with a validation set of $100$ train instances (without augmentation) per class. We then compute scores on the CIFAR10 test set of $10,000$ samples, and achieve $100\%$ accuracy.
\\

In all tasks we assume we can determine at test time the ambiguity of which probably good optimal
scale factor to use, the practical use is addressed later in the section.
\\

This means Learning with Signatures has the capacity to achieve the current state-of-the-art on MNIST and CIFAR10 (assuming you can ascertain the scale factors at test time; which at this point is certainly a very hard assumption), beating all other contenders using only $100$ train instances and $1000$ validation samples to tune the weights. This shows the framework is potentially very good at generalizing from unseen observations.
\\

Classification is performed on the CPU with minimal memory overhead and computing optimal $\lambda_{*}$ by the use of Definition \ref{dfn:lambda} involves only a matrix-vector multiplication.
\\

We have shown very good generalization capabilities of the framework given that we can determine at test time which optimal scale factor to use. 
\\

In practice though, one way to address the problem is to work with fixed $\lambda_{*}$ properly tuned by grid search on the validation set that achieve good accuracy on a given class. Then proceed using one-vs-all, having $n$-binary classifiers that will be used to identify an unseen test sample. Adequate adjustment of the number of signatures used to compute representatives is necessary as well as selection of the order of the truncated signature.
\\

Another approach is to determine at test time the ambiguity of the optimal scale factor to use by finding the solution of the geometrical constraint that Definition \ref{dfn:lambda} implies on the derivation of the weights. See Section \ref{sn:optimization} for a discourse on the procedure.
\\

Finally, an alternative would be to obviate the scale factors and use RMSE and MAE Signature and log-signature as score function by comparison with an element-wise augmented mean of the sample, instead of the given test instance, as proposed in Definition \ref{dfn:classification}. The reason behind this is that comparison between means gives better statistical significance \cite{decurto2022}.

\section{Formulation for Obtaining Scale Factors by Optimization}
\label{sn:optimization}

In this section, we show a general mathematical formulation to determine suitable scale factors by optimization as introduced in Definition \ref{dfn:classification}. These scale factors are modeling the fact that for a fair comparison between an element-wise mean representative and a given new sample of the class, we need a scale factor that calibrates the representative minimizing the distance between the representative of the same class and maximizing the distances between other class representatives. This necessity arises because the score function proposed in \cite{decurto2022} works well for batch comparison between element wise-mean sets but not for a representative computed with several instances and only a given sample. That is, scale factors (which can be in the more general form point-wise masks) add statistical robustness to the comparison.
\\

Let $\lambda_{z}$, \(z =\{0,1,\ldots,m\}\) where $z$ is the corresponding class label, be the scalar factors we want to estimate for the purpose of classification, then the problem can be presented as a concomitant optimization of an objective that is convex, see Equation \ref{eqn:convex}, and an objective that is non-convex, see Equation \ref{eqn:nonconvex}.
\begin{eqnarray}
\minimize_{\lambda_{z}} f(\lambda_{z}) & = & \minimize_{\lambda_{z}} \sum_{z}    \textnormal{MAE}\left(\lambda_{z}\odot\left\{x^{(c)}_{z}\right\}^{n}_{c=1},\left\{y^{(c)}_{z'}\right\}^{n}_{c=1}\right), \label{eqn:convex} \\
\maximize_{\lambda_{z}} g(\lambda_{z}) & = & \maximize_{\lambda_{z}} \sum_{z} \sum_{l \neq z}    \textnormal{MAE}\left(\lambda_{z}\odot\left\{x^{(c)}_{z}\right\}^{n}_{c=1},\left\{y^{(c)}_{l'}\right\}^{n}_{c=1}\right). \label{eqn:nonconvex}
\end{eqnarray}
where $z'$, $l'$ denote a given validation sample from class $z$ and $l$, respectively. Proper constraints on $\lambda$ such as $||\lambda||_{\infty} \leq 1$ can make the problem more tractable.
\\

Definition \ref{dfn:lambda} is an equivalent problem with equality to the formulation that is convex in Equation \ref{eqn:convex}. Utilizing simultaneously the non-convex objective in Equation \ref{eqn:nonconvex} though can lead to a practical implementation that could be used at test time.
\\

To that effect, we can leverage methods from numerical optimization to determine in an iterative way a solution to the defined problem; for instance, BFGS or COBYLA are good candidates. Another approach that is feasible is the Alternating Direction Method of Multipliers (ADMM) \cite{boyd11,magnusson14}, where the method can work in subproblems and determine one objective by convexity and the other one iteratively.

\section{Determining the Problem by an Array of Bandpass Filters}
\label{sn:filters}

An approach closer to Signal Processing, with the perspective of considering the signal as a set of multiple frequency bands \cite{Verdu1986,Berrou1996,Raleigh1998}, where the dominant components characterize uniquely the class we want to determine can be employed to ascertain at test time the ambiguity of which scale factor to use.
\\

Under this assumption, we perform an initial exploration to motivate the techniques that could be put in practice.
\\

For the purpose of evaluation, we can analyze the image representatives using a low pass digital filter Savitzky–Golay. This type of digital filter smooths the input data so that we increase our ability to detect correctly among signal representatives. It uses a corresponding sliding window where polynomial interpolation is carried out through linear least squares. It is particularly good at preserving the tendency of the signal and removing high-frequency components.

\begin{figure}[ht]
\centering
\includegraphics[scale=0.4]{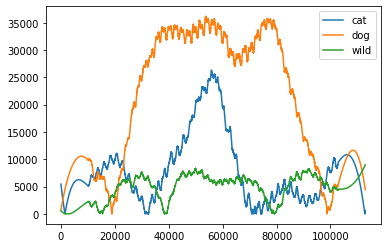}
\caption{SAVITZKY-GOLAY filter with polynomial interpolation of order $3$ and sliding window of size $20001$ in absolute value. Image representatives of AFHQ are computed using a truncated signature of order $3$ and averaged on $1000$ samples.}
\label{fge:00_decurto_and_dezarza}
\end{figure}

In Figure \ref{fge:00_decurto_and_dezarza}, we can see how each AFHQ class representative is distinct from its counterparts. For example, class `dog' presents a significant main lobe on its spectrum, as well as `cat' although with unequal crossing with the abscissa. Class `wild' is more flat with no dominant lobes and also unequal first crossing with the abscissa.

In a way, non-essential information from a given sample, as can visually be seen in Section \ref{sn:explainability}; Figure \ref{fge:02_decurto_and_dezarza}, can be interpreted as noise that corrupts the signal when traveling through a channel of communications. Then, a soft decision mechanism as introduced in \cite{Berrou1996} on a stream of data could design a receiver capable of such a task that achieves optimal capacity of the channel.

\section{Explainability and Data Visualization}
\label{sn:explainability}

Being able to understand model predictions is central to any learning framework. In this section we visualize and analyze the positional encodings behind the use of the Signature Transform. 
\\

The main distinction between the Signature Transform and classical techniques such as Fourier \cite{DuhamelandVetterli90}, Hadamard \cite{Vetterli83,DeCurto22_3} or wavelets \cite{Mallat89,RioulandVetterli91,Vetterli92,Mallat2012,Bruna2013}, is that Signatures are a more general representation that provide a basis for functions on the space of curves \cite{Bonnier2019}.
\\

Bringing into play the identity of Chen \cite{Chen58}, the signature of a stream is straightforward to compute explicitly.
\\

\begin{proposition} \label{prop:chenefficient}
	Let $\mathbf x = (x_1, \ldots, x_n)\in \mathcal S(\mathbb R^d)$ be a stream of data. Then,
	\begin{equation*}
	\mathrm{S}(\mathbf x) = \exp(x_2 - x_1) \otimes \exp(x_3 - x_2) \otimes \cdots \otimes \exp(x_n - x_{n-1}),
	\end{equation*}
	where
	\[\exp(x) = \left(\frac{x^{\otimes k}}{k!}\right)_{k \geq 0} \in T((\mathbb R^d)).\]
\end{proposition}

\begin{figure}[ht]
\begin{subfigure}{.5\textwidth}
\centering
\includegraphics[scale=0.17]{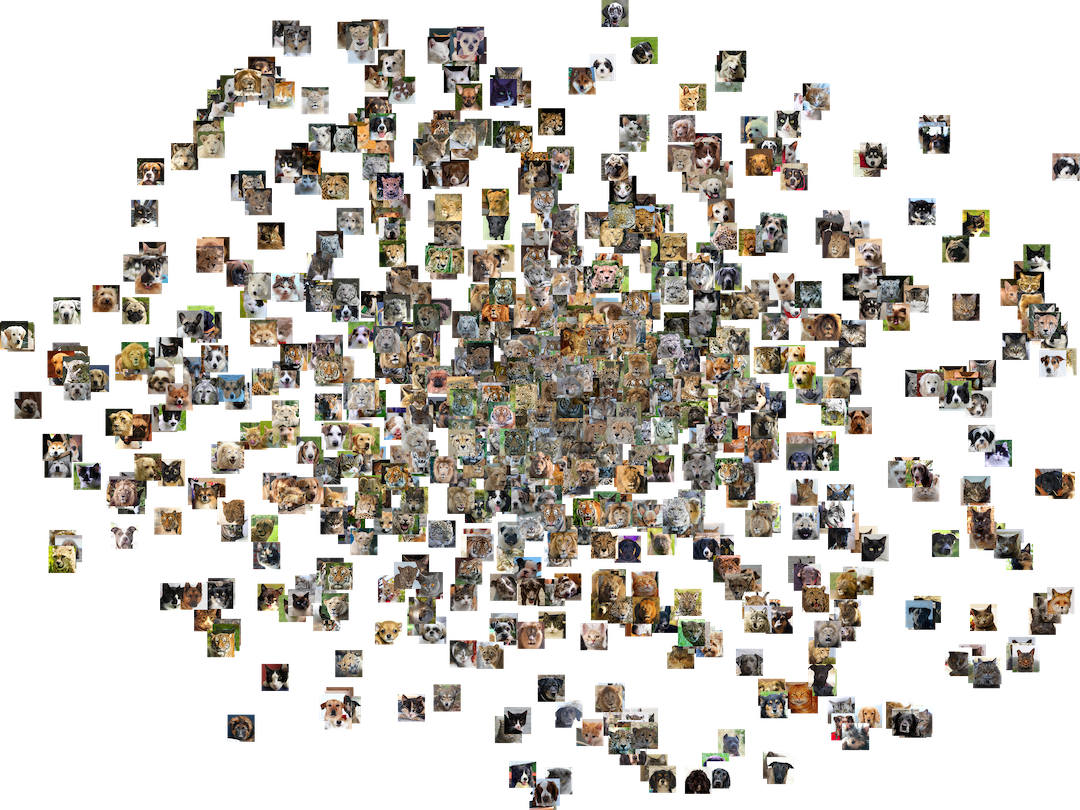}
\end{subfigure}
\begin{subfigure}{.5\textwidth}
\centering
\includegraphics[scale=0.17]{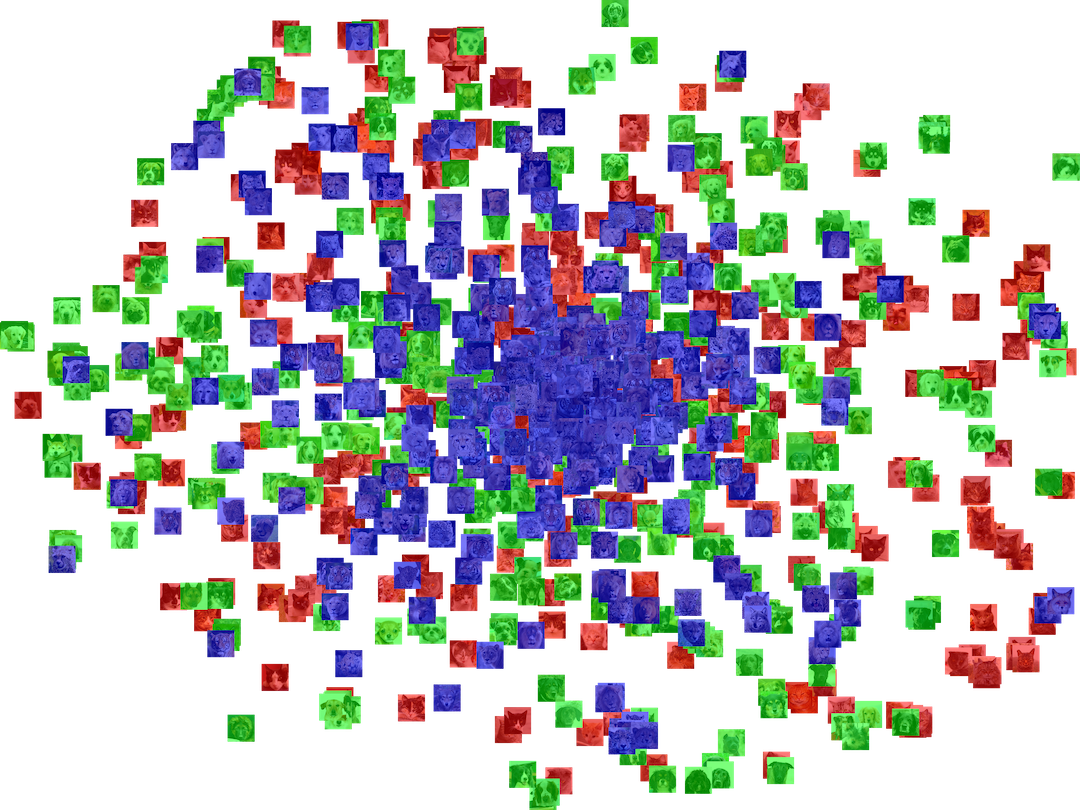}
\end{subfigure}
\caption{PCA Adaptive t-SNE with signatures on 300 images from AFHQ with classes: `cat' (red), `dog' (green) and `wild' (blue).}
\label{fge:01_decurto_and_dezarza}
\end{figure}

We project in the 2D plane by the use of PCA Adaptive t-SNE \cite{decurto2022} the signatures of a subset of 300 samples from AFHQ \cite{Choi2020}, see Figure \ref{fge:01_decurto_and_dezarza}. Signatures are highly multidimensional tensors and by projecting the positional representation into a human understandable 2D plane, we can visually understand the difficulty of the classification task. That being the case, RMSE and MAE Signature and log-signature accomplish a high-dimensional classification task in the spectral domain of the transform. 

Signatures backprojected into the 2D plane appear in concentric circles around a common origin, which makes sense due to the fact that are computed as a tensorial product of exponentials in a high dimensional space.
\\

\begin{figure}[ht]
\begin{subfigure}{.24\textwidth}
\centering
\includegraphics[scale=0.23]{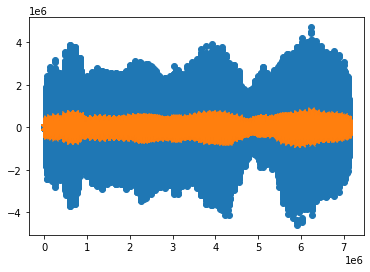}
\caption{}
\end{subfigure}
\begin{subfigure}{.24\textwidth}
\centering
\includegraphics[scale=0.23]{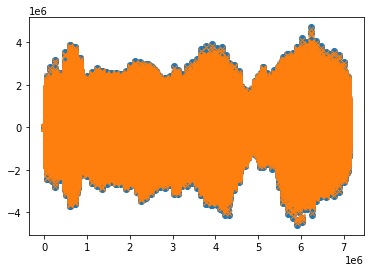}
\caption{}
\end{subfigure}
\begin{subfigure}{.24\textwidth}
\centering
\includegraphics[scale=0.23]{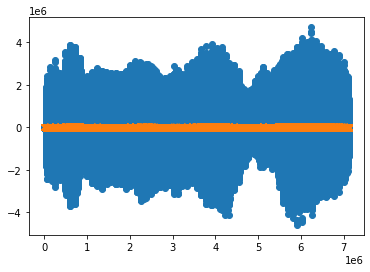}
\caption{}
\end{subfigure}
\begin{subfigure}{.24\textwidth}
\centering
\includegraphics[scale=0.23]{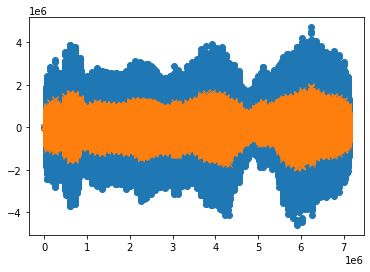}
\caption{}
\end{subfigure}
\caption{Comparison of the spectrum of the Signature Transform for a given AFQH sample (blue `o') and its corresponding transformation (orange `x') with random contrast (a)--(d).}
\label{fge:02_decurto_and_dezarza}
\end{figure}

Comparison at RMSE and MAE Signature and log-signature level is performed between an element-wise mean class representative and a given element-wise mean augmented test sample. To investigate further the effect of the multiplicity introduced by averaging several transformed versions of the same test instance in Figure \ref{fge:02_decurto_and_dezarza} we show the visual comparison of the spectrum using a particular augmentation technique (e.g. random contrast). We can see how slight variations of contrast produce drastic changes of scale (but mild in shape) in the corresponding signatures. This observation is key to understand both the need to put forward an optional augmented element-wise mean for comparison, and also the necessity to choose adequate multiplicative factors $\lambda_{*}$.

Additionally, a principled way to address these issues and mitigate the effects is through appropriate normalization of the sample representatives and instances on validation and test. 

\begin{figure}[ht]
\begin{subfigure}{.3\textwidth}
\centering
\includegraphics[scale=0.23]{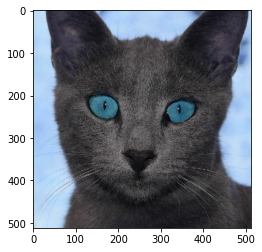}
\caption{}
\end{subfigure}
\begin{subfigure}{.3\textwidth}
\centering
\includegraphics[scale=0.23]{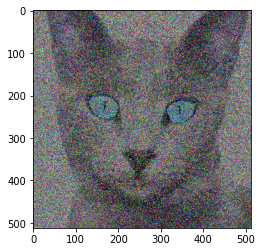}
\caption{}
\end{subfigure}
\begin{subfigure}{.3\textwidth}
\centering
\includegraphics[scale=0.23]{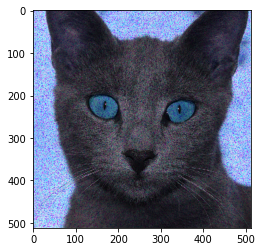}
\caption{}
\end{subfigure}
\begin{subfigure}{.5\textwidth}
\centering
\includegraphics[scale=0.23]{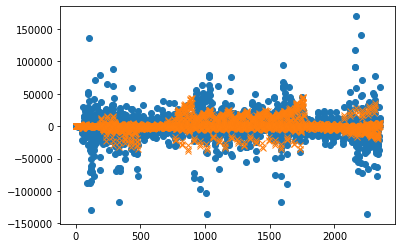}
\caption{}
\end{subfigure}
\begin{subfigure}{.5\textwidth}
\centering
\includegraphics[scale=0.23]{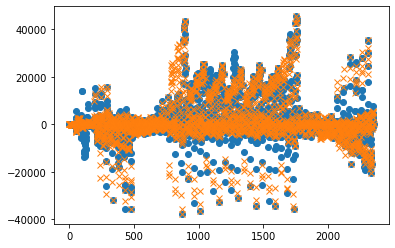}
\caption{}
\end{subfigure}
\caption{Comparison of the spectrum of the Signature Transform for a given AFQH sample (a; orange `o') and its corresponding transformation with noise: speckle (b,d; blue `x'); salt \& pepper (c,e; blue `x').}
\label{fge:03_decurto_and_dezarza}
\end{figure}

To give further elucidation, in Figure \ref{fge:03_decurto_and_dezarza} we show the visual comparison of the spectrum given a sample affected by noise (speckle; salt \& pepper) and see how it modifies the spectrum of the Signature Transform in comparison with the original sample.

\section{Conclusions}
In this manuscript we present a learning framework based on the Signature Transform that potentially achieves state-of-the-art performance with the use of few labels without the need of backpropagation. Capital to the approach is the use of RMSE and MAE Signature and log-signature as score functions for an element-wise mean comparison. A closed-form solution for the weights, scilicet scale factors, that is optimal in probability, is proposed; together with a formulation to obtain them by optimization. Under these premises computation is efficient, fast and robust. These networks can be deployed in CPU, allowing for a fast computation in a resource constrained environment such as in embedded systems, autonomous robots and drones. Techniques of Signal Processing are addressed to further characterize the problem. Perfect accuracy on three difficult benchmarks is reported under the hypothesis that we can correctly determine which $\lambda_{*}$ to use in test. 
%%%%%%%%%%%%%%%%%%%%%%%%%%%%%%%%%%%%%%%%%%%%%%%%%%%%%%%%%%%%

\section*{Acknowledgements}
This work is supported by HK Innovation and Technology Commission (InnoHK Project CIMDA) and HK Research Grants Council (Project CityU 11204821).

%\clearpage

\bibliographystyle{ieeetr}

\end{document}